# Development and Evaluation of a Retrieval-Augmented Generation Tool for Creating SAPPhIRE Models of Artificial Systems

Anubhab Majumder[1], Kausik Bhattacharya[1], and Amaresh Chakrabarti[1]

[1]Department of Design and Manufacturing, Indian Institute of Science, Bengaluru, India
anubhabm@iisc.ac.in

**Abstract.** Representing systems using the SAPPhIRE causality model is found useful in supporting design-by-analogy. However, creating a SAPPhIRE model of artificial or biological systems is an effort-intensive process that requires human experts to source technical knowledge from multiple technical documents regarding how the system works. This research investigates how to leverage Large Language Models (LLMs) in creating structured descriptions of systems using the SAPPhIRE model of causality. This paper, the second part of the two-part research, presents a new Retrieval-Augmented Generation (RAG) tool for generating information related to SAPPhIRE constructs of artificial systems and reports the results from a preliminary evaluation of the tool's success - focusing on the factual accuracy and reliability of outcomes.

## 1 Introduction

### 1.1 Motivation

Historically, design-by-analogy (DbA) played a crucial role in driving design creativity [1]. The DbA method uses analogical reasoning to draw inspiration from a source domain to solve a design problem in the target domain [2]. Supporting the DbA method requires four primary steps: encoding, retrieval, mapping, and evaluation [2,3]. Encoding refers to storing and organising information about artificial or biological entities that belong to the source domain. Retrieving involves searching the entities from the source domain relevant to the design problem in the target domain. The source entities and the target problem are mapped by identifying their similarities to determine the analogies that solve the design problem. The final step is evaluating the generated ideas based on specific criteria such as fluency, novelty, usefulness, etc [4]. When it comes to encoding and mapping, using a structured ontology to represent source entities or stimuli helps designers follow the DbA process efficiently [5]. In our current work, we aim to support DbA by leveraging Large Language Models (LLMs) to automatically generate descriptions of source entities or stimuli, using a causal ontology called the SAPPhIRE model of causality [6].

### 1.2 Research Plan

We plan our research activities following the 'Design Research Methodology' (DRM) framework [7]. The research activities associated with the different stages of DRM are summarised in **Fig. 1**.

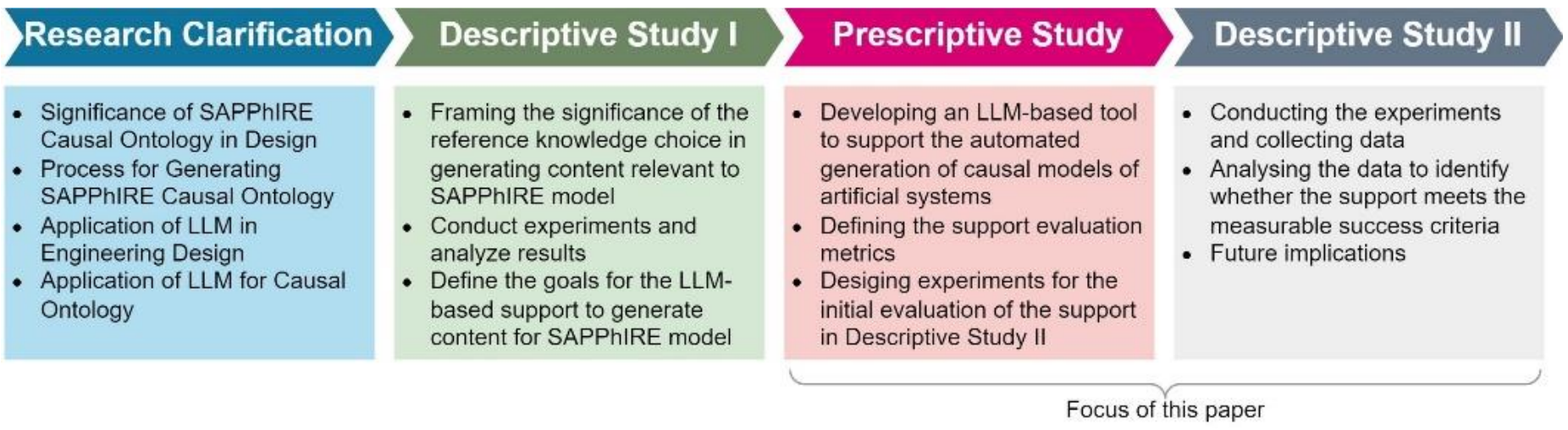

**Fig. 1.** DRM [12] stages and associated research activities.

The research clarification (RC) activity primarily involves reviewing existing literature and clarifying the research aims, context, and scope. In Section 2, we have briefly addressed the current state-of-the-art related to DbA and the SAPPhIRE model's role in supporting DbA. A more detailed discussion of the existing literature is available in [8]. As an outcome of RC, we have formulated the overarching aim of this work: *"to leverage Large Language Models (LLMs) in creating a large database of analogues that includes structured descriptions of systems using the SAPPhIRE model of causality."* Though our long-term plan is to create databases that include artificial and biological systems, we limit our current scope to artificial or engineering systems only.

In Descriptive Study I (DS-I), as reported in [9], we investigated whether LLMs could generate the content relevant to a SAPPhIRE representation of artificial systems instead of manually searching and extracting information from multiple technical documents. We also empirically investigated the significance of the reference knowledge provided in the LLM prompt for content generation. Providing reference knowledge or context to the LLM while generating a response is known as Retrieval-Augmented-Generation (RAG). DS-I focused specifically on generating content that provides the condition(s) of a physical interaction inside a system, and we found RAG to be a promising approach for this purpose.

To achieve our overarching aim, we need to create a robust tool as a part of the prescriptive study (PS) that can support the generation of SAPPhIRE models for a given system. The overall success of this work is defined in terms of the following criteria: (a) Factual accuracy: The SAPPhIRE models generated by the tool must adhere to the definitions of SAPPhIRE constructs, and the corresponding information generated by the tool must be supported by a reference knowledge source; (b) Reliability: The tool must support the aforementioned criteria on successive trials involving different artificial or biological systems. In Descriptive Study II (DS-II), we will evaluate the proposed tool based on these success criteria.

### 1.3 Objectives

This paper reports the findings from the Prescriptive Study (PS) and initial Descriptive Study II (DS-II), as outlined in **Fig. 1**. The key objectives of this paper are formulated as follows:

1. To introduce a new Retrieval-Augmented Generation (RAG) tool designed for generating information related to SAPPhIRE constructs of artificial systems.
2. To present results from a preliminary evaluation of the tool's success - focusing on its factual accuracy and reliability - as defined in the success criteria.

## 2 Related Work

Researchers have developed several computational supports [10-12] for DbA that provide hand-curated databases of analogues represented using different structured ontology models. AskNature[1] provides an open-source database of biological inspirations. The database is organised utilising the biomimicry taxonomy, which categorises biological information based on functions [10]. DANE (Design-by-Analogy to Nature Engine) uses the Structure-Behaviour-Function (SBF) based ontology to store information about existing artificial and biological systems [11]. Chakrabarti et al. [6] developed IDEA-INSPIRE, which uses a causal ontology model called SAPPhIRE and enables analogy retrieval from artificial and biological domains.

The SAPPhIRE model has seven elementary constructs or abstraction levels [6]: State-changes, Actions, Parts, Phenomena, Inputs, oRgans, and Effects, as shown in **Fig. 2**(a). These constructs and their relationships provide a causal explanation of how a system achieves its intended function. **Fig. 2**(b) shows an example where the SAPPhIRE model explains how a hot body cools down in the presence of a surrounding fluid medium. Past research [13,14] demonstrated the potential of the SAPPhIRE model in representing complex analogues and found it to be more comprehensive and flexible compared to other existing models, such as SBF, for systems with varying levels of complexity. Researchers [4] have also conducted experiments with engineering designers and found that the SAPPhIRE-based DbA approach helps the designers generate novel ideas and supports developing those ideas into realisable prototypes. Notably, the Indian Space Research Organization (ISRO) employed IDEA-INSPIRE to generate innovative products for space applications [4].

[1] https://asknature.org/

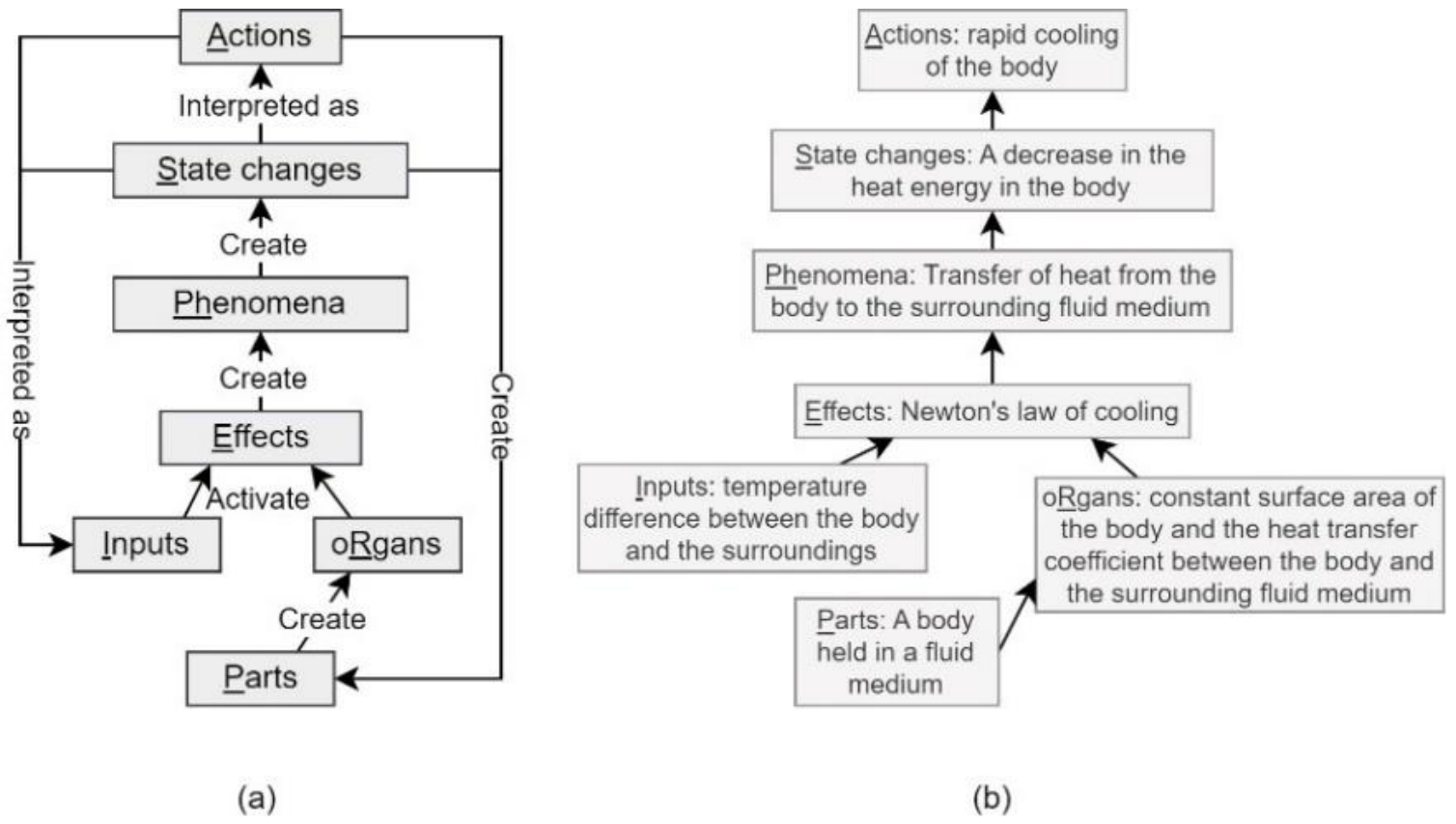

**Fig. 2.** (a) The SAPPhIRE model of causality (adapted from [6]), (b) An example SAPPhIRE model explaining how a hot body cools down [9].

However, creating SAPPhIRE models of existing systems or analogues is an effort-intensive process that requires human experts to source necessary details related to a system from multiple documents [8]. The dependency on human experts to generate SAPPhIRE models restricts us from scaling the IDEA-INSPIRE database and, therefore, limits its use in design practice. However, recent advancements in AI, particularly Large Language Models (LLMs), offer promising opportunities for automating the creation of large datasets of analogues, potentially overcoming these limitations. This paper reports our initial attempt to use AI to create SAPPhIRE models of artificial systems.

# 3 Implementation

## 3.1 Tool Overview

The framework outlined in **Fig. 3** captures the overall workflow of the proposed tool. Given the name of a system $X$ as an input, the purpose of the tool is to generate an output $Y = \{y_i\}_{i=1}^{7}$ that contains the list of seven SAPPhIRE constructs corresponding to the input system $X$ and the generated output $Y$ should be grounded in a knowledge source document(s) $D$. We decompose this task into four sub-tasks: knowledge source embedding, hypothetical document embedding, context retrieval, and generating corrected responses. In this implementation, we have used Wikipedia articles as $D$ due to its comprehensive coverage and the availability of an open-source API. However, it is important to note that the framework is not dependent on Wikipedia specifically—any structured or unstructured knowledge source can be used as $D$.

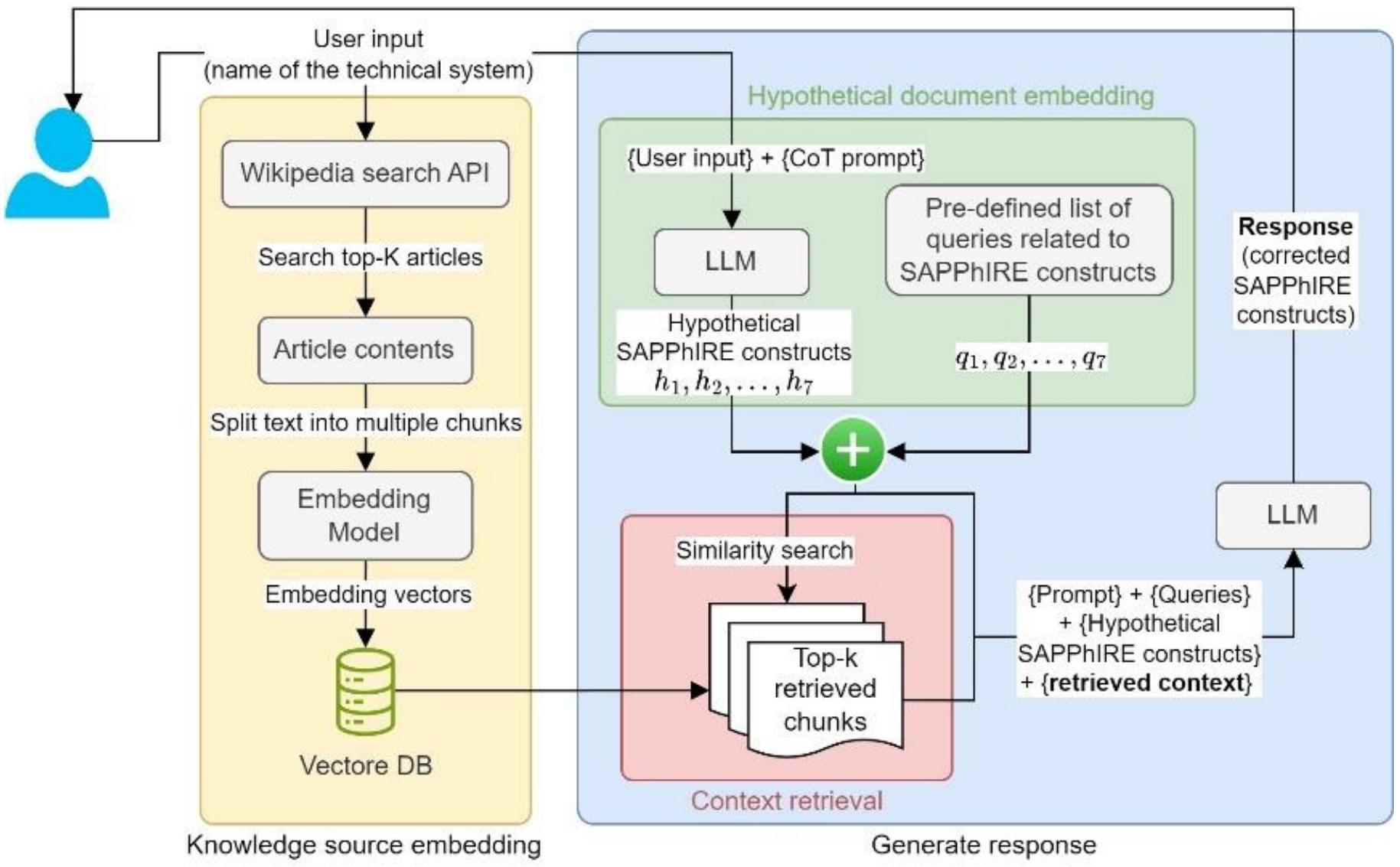

**Fig. 3.** Workflow of our proposed Tool.

**Knowledge Source Embedding.** First, we retrieve relevant (top $K$) knowledge source documents, $D = \{d_i\}_{i=1}^{K}$, for the given user input $X$ using Wikipedia API[2]. Here, $K$ is a pre-defined number which denotes the maximum number of Wikipedia articles we want to use as knowledge sources for the given input $X$. Then, all the text contents are extracted from $d_i$s and merged into a single document. At this point, $D$ contains a large number of tokens (~ words). Considering the LLM's maximum context length and cost, adding the whole text content into the prompt for LLM response generation is not always possible. Hence, a smarter approach is to split the merged text content of $D$ into multiple chunks of text and then provide the LLM only the relevant pieces of information while generating the output $Y$. A *RecursiveCharacterTextSplitter* is deployed to split the text into several chunks based on a designated set of characters (e.g., "\n\n", "\n", etc.). Now, we have a corpus $C = \{c_i\}_{i=1}^{l}$ where $c_i$ denotes the $i^{th}$ chunk of the corpus. Finally, we use an embedding model, *OpenAIEmbeddings*, to map the text chunks to their respective high-dimensional vector representations and store them in a vector database. We utilise ChromaDB[3] to store and search vector data.

**Hypothetical Document Embedding.** The next step is to retrieve the relevant chunks from the vector database for a given query related to a particular SAPPhIRE construct of the system $X$. For this task, a generic list of pre-defined queries $Q = \{q_i\}_{i=1}^{7}$ is prepared where each query asks a question related to a particular SAPPhIRE abstraction level. The list of pre-defined queries is shown in **Table 1**. Now, for retrieving relevant chunks, one straightforward approach could be creating a vector representation of a query $q_i$ and get the top $k$ number of chunks by performing a similarity search (e.g., cosine similarity) in the vector database of corpus $C$. However, researchers found that, in many cases, this approach often misses out on meaningful chunks essential for generating better responses [15]. Formulating a precise and clear query is a difficult task, and queries that are not appropriately optimised result in subpar retrieval effectiveness. In our case, the query $q_i$ is complex as it involves terminologies related to the SAPPhIRE model that are not generally used in Wikipedia articles. For example, the term 'organ' is a specialised vocabulary used in SAPPhIRE, but it may exist in a text chunk, conveying a very different meaning. Hence, relying on a straightforward query search approach may not be effective in our case. Researchers have proposed various strategies to handle this issue [15], such as query expansion and transformation. In our proposed framework, we have incorporated one of the query transformation methods named 'Hypothetical Document Embedding' (or HyDE) [16] to improve the retrieval process. The HyDE approach uses an LLM to generate a hypothetical document, $H = \{h_i\}_{i=1}^{7}$, where, $h_i$ as a presumed response to the original query $q_i$. Here, we use a chain-of-thought prompting technique [17] to generate the $h_i$s. The prompting technique is discussed in detail in the following section. We formalise this step as follows:

$$H = \{h_i\}_{i=1}^{7} = f(\tau(X)) \quad (1)$$

In **Eqn. 1**, $\tau$ denotes the prompt template and $f$ is the response generation function.

**Table 1.** List of pre-defined queries related to SAPPhIRE constructs

| Abstraction level $i$ | Query $q_i$ |
|---|---|
| Actions | *What are the Actions or overall purpose served by the system?* |
| Effects | *What are the scientific laws, principles, or theories that govern the system's operation?* |
| Inputs | *What inputs are necessary for the system to operate?* |
| oRgans | *What are the properties and conditions that remain constant during the interaction between the system and its environment and are necessary for activating the operation of the system?* |
| Parts | *What are the physical components and interfaces that constitute the system and its environment.?* |
| Phenomena | *What physical phenomena occur during the operation of the system?* |
| State changes | *What state changes occur in the system during its operation? Mention the initial and final states of the properties involved.* |

**Context Retrieval.** After generating $h_i$, the query $q_i$ is transformed into $q_i'$ where $q_i' = [q_i + h_i]$. Here, '+' represents a concatenation operation. We use $q_i'$ to perform a search operation in the vector database and retrieve the top k chunks, $r_i = \{c_j\}_{j=1}^{k}$, $r_i \subset C$, from the knowledge source corpus $C$. Note that the hypothetical document, i.e., $h_i$, generated during this process may not be factually correct or 'real.' However, researchers [16] have shown that, by following the HyDE approach, we no longer rely on embedding similarity from a document to a query but seek document-to-document embedding similarity, resulting in superior retrieval results compared to using the query itself.

[2] https://pypi.org/project/Wikipedia-API/
[3] https://pypi.org/project/chromadb/

**Generating Corrected Responses.** At this juncture, for each query $q_i$, we have a corresponding hypothetical response $h_i$ (might be factually incorrect) and a corresponding set of relevant contexts, $r_i$, retrieved from the knowledge source (i.e., the Wikipedia articles). The final task is to correct the $h_i$ based on the given context, $r_i$, and generate the corrected response $y_i$ using another LLM. For this task, we create another prompt template $\rho$ and define a 'corrected response generation' function $g$ as follows:

$$y_i = g(\ \rho(X, q_i, h_i, r_i)) \tag{2}$$

## 3.2 Prompt Engineering

Prompt engineering involves designing and refining instructions for the LLM to generate the desired output [18]. Our proposed framework requires two different prompting strategies, $\tau$ and $\rho$, as shown in **Eqn. 1** and **Eqn. 2**, respectively.

For hypothetical SAPPhIRE construct generation, we employ a prompt template $\tau$, by taking inspiration from the chain-of-thought (CoT) [17] prompting strategy. **Fig. 4** describes the overall structure of the prompt $\tau$. It takes the system name $X$ as the only input variable. Here, we first introduce SAPPhIRE to the LLM, followed by instructions for the LLM to generate appropriate knowledge related to the SAPPhIRE constructs. The instructions include a series of intermediate reasoning steps that are expected to help the LLM to 'think' about different aspects of the given system $X$ before jumping into a final answer. For example, to generate knowledge about Parts, the reasoning step is described as follows: "*Generate knowledge about the `Parts` by analysing all possible components and interfaces that constitute the system and its environment. Parts are generally `Material Nouns` representing material entities of the system and its environment*." The prompt ends with a set of format instructions that helps to parse the LLM-generated outputs (i.e., $h_i$) into a suitable format.

**Fig. 5** describes the overall structure of the prompt template $\rho$ which is used for the corrected response generation. As shown in **Eqn. 2,** this prompt template has four input variables: system name ($X$), query ($q_i$), the corresponding hypothetical SAPPhIRE construct ($h_i$), and the retrieved context ($r_i$). As shown in **Fig. 5**, here, we explicitly instruct the LLM to keep the output grounded in the facts of the given context. Unlike conventional RAG prompts, we instruct the model to correct the existing hypothetical answer instead of synthesising a new response from scratch.

```
You are a helpful assistant. Your job is to create knowledge regarding any given artificial or natural system description by using an ontology named `SAPPhIRE`.
The acronym `SAPPhIRE` stands for the following constructs: `State changes` (S), `Actions` (A), `Parts` (P), `physical Phenomena` (Ph), `Inputs` (I), `oRgans` (R), `physical Effects` (E).
###
Following are the definitions of the `SAPPhIRE` constructs:
{Definitions of the SAPPhIRE constructs}
###
A brief explanation of the workings of these constructs is given below:
`Parts` are necessary for creating `oRgans`. `oRgans` and `Inputs` are necessary for activation of `physical Effects`, which in turn is necessary for creating `physical Phenomena` and `State changes`. `State changes` are interpreted as `Actions`, which describe the overall purpose of the system.
###
Important Note: Here, all the `physical Phenomena` should be governed by corresponding `physical Effects` which are valid scientific laws, principles, or theories from engineering science or natural science domain.
###
You are given the following `System`:
{System name}
###
Think step by step to generate appropriate knowledge that will be helpful to create the `SAPPhIRE` constructs of the given `System`:
{Reasoning steps}
###
{Output format instructions}
```

**Fig. 4.** The prompt template used for hypothetical document generation.

```
You are a helpful assistant. Your task is to provide relevant information about any specified `SAPPhIRE` ontology construct related to a given artificial or natural system. You must use only the information provided to you in the `Context` section. Ensure that your responses are accurate, concise, and directly relevant to the query.
###
The acronym `SAPPhIRE` stands for the following constructs: `State changes` (S), `Actions` (A), `Parts` (P), `physical Phenomena` (Ph), `Inputs` (I), `oRgans` (R), `physical Effects` (E).
###
Following are the definitions of the `SAPPhIRE` constructs:
{Definitions of the SAPPhIRE constructs}
###
The relationships between these constructs are given below:
`Parts` are necessary for creating `oRgans`. `oRgans` and `Inputs` are necessary for activation of `physical Effects`, which in turn is necessary for creating `physical Phenomena` and `State changes`. `State changes` are interpreted as `Actions`, which describe the overall purpose of the system.
###
Important Note: Here, all the `physical Phenomena` should be governed by corresponding `physical Effects` which are valid scientific laws, principles, or theories from engineering science or natural science domain.
###
Your users are asking you to correct the `Hypothetical Answer` that was generated for a paticular `Query` related to a particular `SAPPhIRE` construct of the given `System`.
You will be shown the `System`, `Query`, `Hypothetical Answer`, and the relevant `Context`. Correct the `Hypothetical Answer` provided by the user. You must use only the information provided to you in the `Context` section. Keep your answer grounded in the facts of the `Context`. Make sure the `Corrected Answer` is as concise as the `Hypothetical Answer`.
###
`System`: {system name} \n\n
`Query`: {query} \n\n
`Hypothetical Answer```: {hypothetical_answer} \n\n
`Context`: {context_str}. \n\n
`Corrected Answer`:
```

**Fig. 5.** The prompt template used for corrected response generation.

# 4 Evaluation

## 4.1 Experimental Setup

We experiment with our proposed tool for three test cases that involve generating SAPPhIRE models of an orifice plate, a thermoelectric cooler (TEC), and a solenoid valve. To split the Wikipedia content into multiple chunks, we arbitrarily set the maximum chunk size to 1024 characters with a chunk overlap of 256 characters.

We use OpenAI's state-of-the-art "GPT-4o" model for hypothetical document generation. Considering traditional benchmarks, specifically GPQA (Graduate-Level Google-Proof Q&A Benchmark) [19], "GPT-4o" performs better than existing LLMs such as "Gemini 1.5 Flash" and "Llama 3.1 405b" [20]. The GPQA consists of multiple-choice questions written by domain experts in biology, physics, and chemistry. The compelling rationale behind our selection of "GPT-4o" for generating hypothetical documents is based on its superior performance in the GPQA benchmark.

Unlike the hypothetical document generation task, the corrected response generation task does not rely much on the LLM's self-knowledge since we provide the required knowledge within the prompt. Therefore, we opt for OpenAI's "GPT-3.5-turbo-0125" model for this task, as it offers a lower cost alternative than the more expensive "GPT-4o" while still reasonably meeting our task requirements. However, it is important to note that our proposed framework supports the integration of any other LLMs as defined by the end user.

The output from an LLM model can also be controlled by tweaking the configuration settings provided by OpenAI. These configurations include *temperature*, *top-p*, *frequency penalty*, *presence penalty*, etc. *Frequency penalty* and *presence penalty* regulate the repetitiveness of words used while generating the texts, while *temperature* and *top-p* govern the randomness of the output. In our experiments, we kept all the default parameters suggested by OpenAI, but we set the *temperature* parameter to 0 to reduce the randomness in the text generation process.

## 4.2 Evaluation Metrics

Since we do not have any ground truth dataset, we use the state-of-the-art automated RAG evaluation tool called the 'RAG Triad' provided by TruLens[4]. This approach emphasises three quality scores of the RAG system: answer relevance, groundedness (or faithfulness), and context relevance [15]. The answer relevance score determines whether the response ($y_i$) generated by the system is pertinent to the original query ($q_i$). The groundedness score determines whether the responses ($y_i$) is supported by the retrieved context ($r_i$). The context rele-vance score ensures that the retrieved context ($r_i$) is relevant to the original query ($q_i$). TruLens utilises LLMs to adjudicate these quality scores. Apart from giving scores, the TruLens dashboard also offers supporting evidence or reasoning behind the scores. In our experiments, we have used TruLens's default LLM provider, "GPT-3.5-turbo."

[4] https://www.trulens.org/trulens_eval/getting_started/core_concepts/rag_triad/

As discussed in Section 2, the success of our proposed framework is defined based on two criteria: factual accuracy and reliability. We consider the three quality scores provided by TruLens as a proxy (also known as measurable success criteria in design research methodology [7]) to the factual accuracy criterion. Due to the non-deterministic nature of LLM outputs, we repeated the experiments ten times for each test case (involving an orifice plate, TEC, and solenoid valve). Then, we observe the change in the average quality scores over these ten successive trials to evaluate the reliability of the proposed RAG tool.

# 5 Results and Discussion

## 5.1 Example of a SAPPhIRE Model Generated by the Tool

**Fig. 6** illustrates two representations of the SAPPhIRE model of a solenoid valve. **Fig. 6**(a) depicts the initial SAPPhIRE model generated by the "GPT-4o" model as part of the hypothetical document embedding process. The initial outcome is promising, as the LLM successfully generates relevant texts for the different SAPPhIRE constructs corresponding to the given system. Moreover, these generated texts align well with the definitions of the SAPPhIRE constructs. The results show the potential of our chain-of-thought prompting strategy (as shown in **Fig. 4**) in helping the LLM understand the SAPPhIRE constructs and generate appropriate outputs.

After generating the hypothetical document, we proceed with our subsequent RAG workflow. **Fig. 6**(b) showcases the final SAPPhIRE model of the solenoid valve, which represents the 'corrected' version obtained from the RAG tool. Despite setting the *temperature* to 0, we observed that the LLM tends to produce varied outputs in successive trials. Therefore, we repeated the RAG workflow ten times to generate the final model and selected outputs with the highest TruLens average quality scores. **Fig. 6**(b) shows the instances where our tool has effectively filtered out information from the hypothetical model that is not present in the retrieved knowledge sources, thereby achieving a higher groundedness score. However, the responses are not entirely error-free. For example, in **Fig. 6**(b), the tool removed "Bernoulli's principle" from the Effects construct as it was not mentioned in the Wikipedia article but ignored removing the related input, 'Pressure differential,' from the Inputs construct. The absence of memory in the prompt $\tau$ (**Fig. 5**) causes this ambiguity. Here, the LLM doesn't know what correction it has done before for other constructs and hence fails to maintain consistency within the constructs while correcting the hypothetical model.

## 5.2 Evaluation Scores

TruLens assigns all three quality evaluation scores on a scale from 0 to 1, where values between 0 and 0.5 indicate low scores, values between 0.8 and 1 indicate high scores, and values between 0.5 and 0.8 represent medium scores. In **Fig. 7**, we plot the three quality scores for each SAPPhIRE construct, where each column represents average scores considering all three test cases. The following observations are made:

- Except for Parts and oRgans, the groundedness scores are high for the rest of the constructs. This result shows that the language model tends to ignore the retrieved context while correcting the descriptions of Parts and oRgans.
- None of the constructs shows mean answer relevance scores below 0.7. This result indicates that even though the groundedness scores are low in cases of Parts and oRgans, the TruLens evaluator LLM still found the generated responses relevant to the queries listed in **Table 1**.
- The average context relevance scores are either high or medium for all the constructs except for oRgan. This result implies the following possible scenarios: (a) the Wikipedia articles related to the test case systems may not contain enough information about oRgans; (b) The context retrieval function may have failed to retrieve relevant chunks of information.

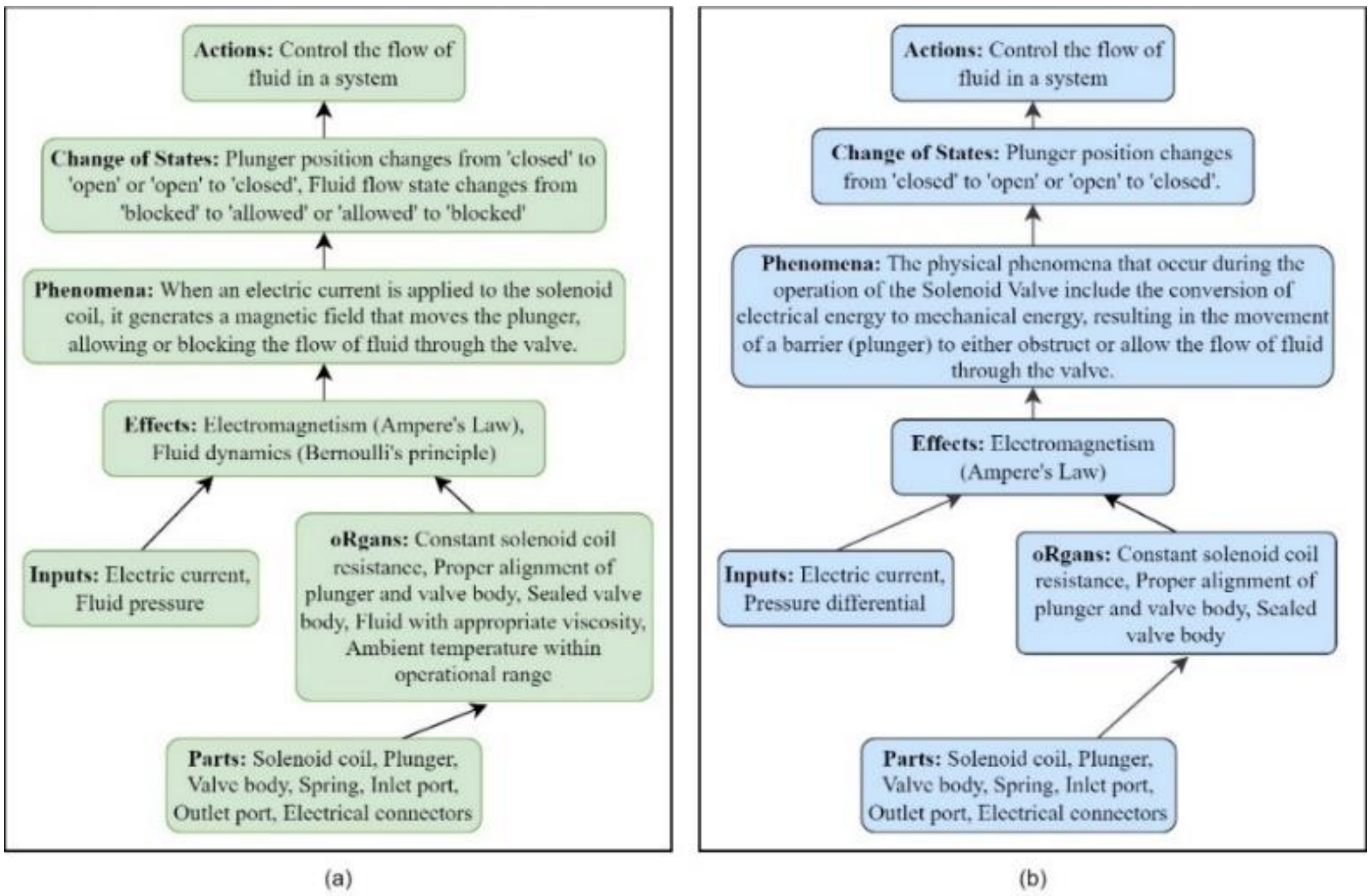

**Fig. 6.** (a) SAPPhIRE model of the solenoid valve generated during hypothetical document embedding, (b) The final SAPPhIRE model of the solenoid valve obtained from the RAG tool.

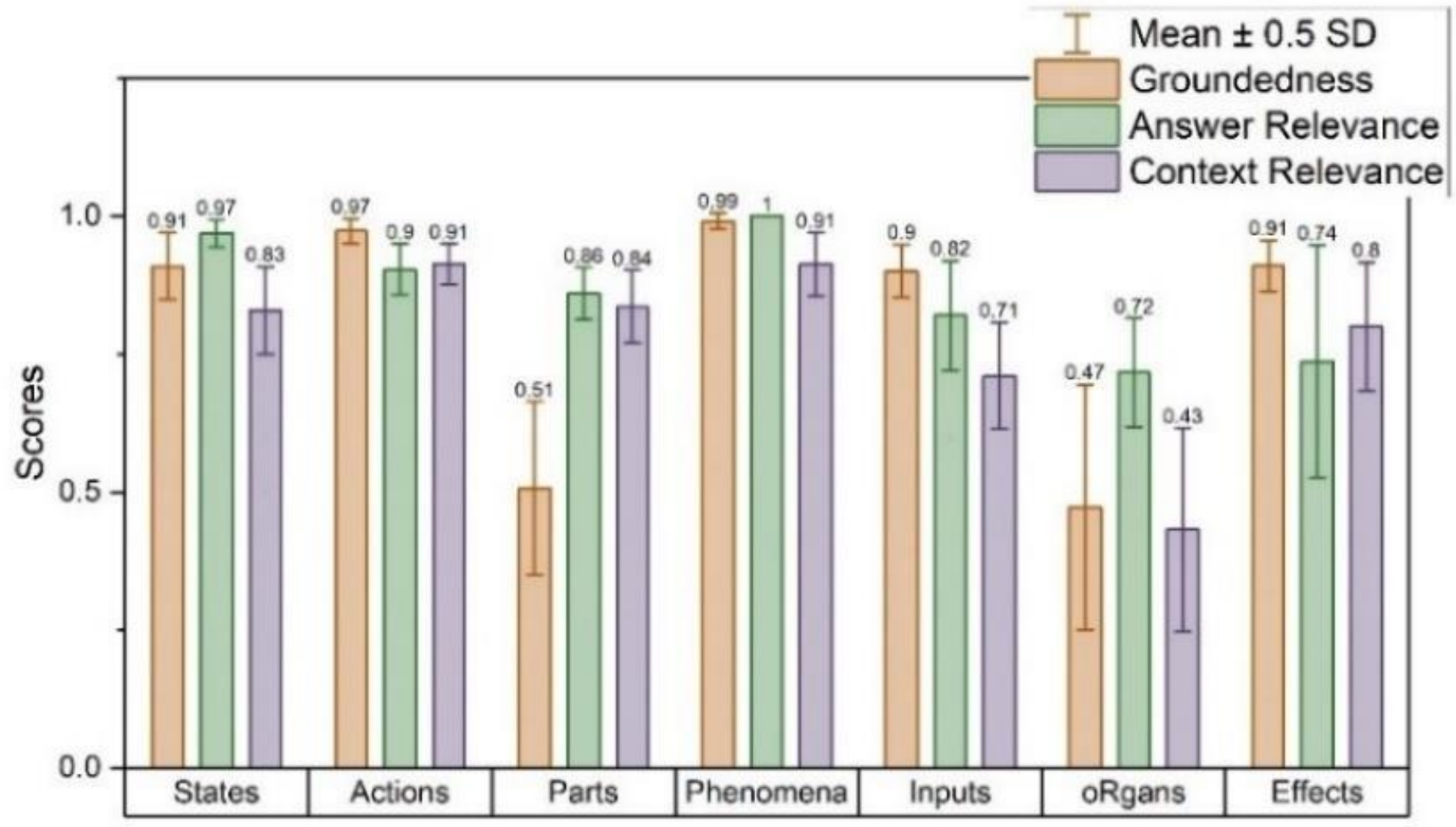

**Fig. 7.** Average scores of groundedness, answer relevance, and context relevance for each SAPPhIRE construct considering the three test cases.

**Table 2** provides the descriptive statistics for the overall quality scores across three test cases. In our experiments, the maximum sample size for each quality score was 210 (7 queries × 10 repetitions × 3 test cases). However, in many cases, the TruLens evaluator failed to provide the scores for groundedness and context relevance due to API call failures, which resulted in a reduced sample size. The statistics for answer relevance score (M= 0.86, 95% CI [0.83, 0.89]) suggest that our proposed tool generated descriptions of SAPPhIRE constructs relevant to the respective SAPPhIRE definitions across all the test cases. As far as the answer relevance is concerned, the proposed tool is found to be highly reliable. However, the groundedness and context relevance statistics (see **Table 2**) depict a lower mean and wider confidence interval range compared to answer relevance, suggesting that the proposed tool is moderately reliable regarding groundedness and context relevance quality.

**Table 2.** Overall quality scores are calculated by considering all the test cases and their respective SAPPhIRE constructs.

| | Quality Scores | | |
|---|---|---|---|
| | Groundedness | Answer Relevance | Context Relevance |
| Sample size | 183 | 210 | 167 |
| Mean | 0.81 | 0.86 | 0.79 |
| Standard Deviation | 0.29 | 0.22 | 0.24 |
| 95% Confidence Interval | [0.77, 0.85] | [0.83, 0.89] | [0.76, 0.83] |

# 6 Conclusion

In this paper, we introduced a novel Retrieval-Augmented-Generation (RAG) tool designed to leverage Large Language Models for generating information pertaining to SAPPhIRE constructs of artificial systems. The tool accepts the system name as input and produces the corresponding SAPPhIRE model. To address the challenge of hallucinations, the tool utilises Wikipedia articles related to the input system to ground its generated responses. We defined two primary success criteria for the tool: factual accuracy and reliability. To evaluate the tool's performance, we conducted a series of experiments involving three test cases: an orifice plate, a thermoelectric cooler (TEC), and a solenoid valve. Factual accuracy was measured using quality scores (judged in terms of answer relevance, context relevance, and groundedness) provided by the RAG-Triad evaluation framework, a readily available LLM-based RAG assessment tool. Reliability was assessed by repeatedly running the evaluation to check the consistency of the quality of outcomes generated by the tool. Our experimental results indicate that our proposed tool has the potential to create accurate and reliable SAPPhIRE models. The grounding of responses in relevant Wikipedia articles significantly mitigates the risk of hallucinations, thereby enhancing the tool's overall utility in automating the process of populating datasets for DbA tools. Though the overall performance of the tool was found to be satisfactory, the tool struggled to ground the response related to oRgans and Parts. This difficulty likely stems from insufficient information in the Wikipedia articles concerning these SAPPhIRE constructs. Consequently, we should carefully consider the comprehensiveness of the input reference knowledge source and the design of prompts to maximise the tool’s effectiveness. Hence, our future work will focus on refining the prompts and the context retrieval approach to address these gaps. Additionally, further experiments will involve an extensive set of test cases and a ground truth dataset prepared by human experts to validate the tool's success.